\def\year{2018}\relax
\documentclass[letterpaper]{article} 
\usepackage{aaai18}  
\usepackage{times}  
\usepackage{helvet}  
\usepackage{courier}  
\usepackage{url}  
\usepackage{graphicx}  
\usepackage{amsmath}
\usepackage{amssymb}
\usepackage{graphicx}
\newcommand{\cev}[1]{\reflectbox{\ensuremath{\vec{\reflectbox{\ensuremath{#1}}}}}}
\usepackage{framed}
\usepackage{subfigure}
\usepackage{booktabs}
\usepackage{empheq}
\usepackage{array}
\usepackage{color,soul}
\definecolor{tableau20}{RGB}{174, 199, 232}
\setstcolor{green}
\sethlcolor{tableau20} 
\usepackage{enumitem}
\newcommand{\unk}{$ \langle \textit{unk} \rangle $}

\newcommand{\ourModel}{SeqCopyNet}
\newcommand{\significant}{$ {}^{\text{\textbf{-}}} $}
\newcommand{\otherpaper}{$ {}^{\ddag} $}

\newcommand{\red}[1]{{\color{red}{#1}}}
\renewcommand{\red}[1]{#1}

\newcommand{\citet}[1]
{\citeauthor{#1}~\shortcite{#1}}
\newcommand{\citep}{\cite}

\begin{document}
%

\title{Sequential Copying Networks}

\author{Qingyu Zhou$^\dag$\thanks{Contribution during internship at Microsoft Research.} \hspace{0.15cm} Nan Yang$^\ddag$ \hspace{0.15cm} Furu Wei$^\ddag$ \hspace{0.15cm} Ming Zhou$^\ddag$ \\
	$^\dag$Harbin Institute of Technology, Harbin, China \\
	$^\ddag$Microsoft Research, Beijing, China \\
	{\tt qyzhgm@gmail.com} \hspace{0.15cm} {\tt \{nanya, fuwei, mingzhou\}@microsoft.com}
}

\maketitle
\begin{abstract}
Copying mechanism shows effectiveness in sequence-to-sequence based neural network models for text generation tasks, such as abstractive sentence summarization and question generation. 
However, existing works on modeling copying or pointing mechanism only considers single word copying from the source sentences. 
In this paper, we propose a novel copying framework, named Sequential Copying Networks (\ourModel{}), which not only learns to copy single words, but also copies sequences from the input sentence.
It leverages the pointer networks to explicitly select a sub-span from the source side to target side, and integrates this sequential copying mechanism to the generation process in the encoder-decoder paradigm.
Experiments on abstractive sentence summarization and question generation tasks show that the proposed \ourModel{} can copy meaningful spans and outperforms the baseline models.
\end{abstract}

\section{Introduction}
Recently, attention-based sequence-to-sequence (seq2seq) framework \cite{sutskever2014sequence,bahdanau2014neural} 
has achieved remarkable progress in text generation tasks, such as abstractive text summarization \cite{rush-chopra-weston:2015:EMNLP}, question generation \cite{zhou2017neural} and conversation response generation \cite{vinyals2015neural}. In this framework, an encoder is employed to read the input sequence and produce a list of vectors,  which are then fed into a decoder to generate the output sequence by making word predictions one by one through the softmax operation over a fixed size target vocabulary.

It has been observed that seq2seq suffers from the unknown or rare words problem \cite{luong-EtAl:2015:ACL-IJCNLP}.
\citet{gulcehre-EtAl:2016:P16-1} and \citet{gu-EtAl:2016:P16-1} makes the key observation that in tasks like summarization and response generation, rare words in the output sequence usually can be found in the input sequence. Based on this observation, they propose a copying mechanism to directly copy words to the output sequence from input, which alleviates the rare word problem. In their work, every output words can be either generated by predicting words in the target vocabulary or copied from the input sequence.

\begin{figure}[htb]
	\centering
	\includegraphics[scale=0.28]{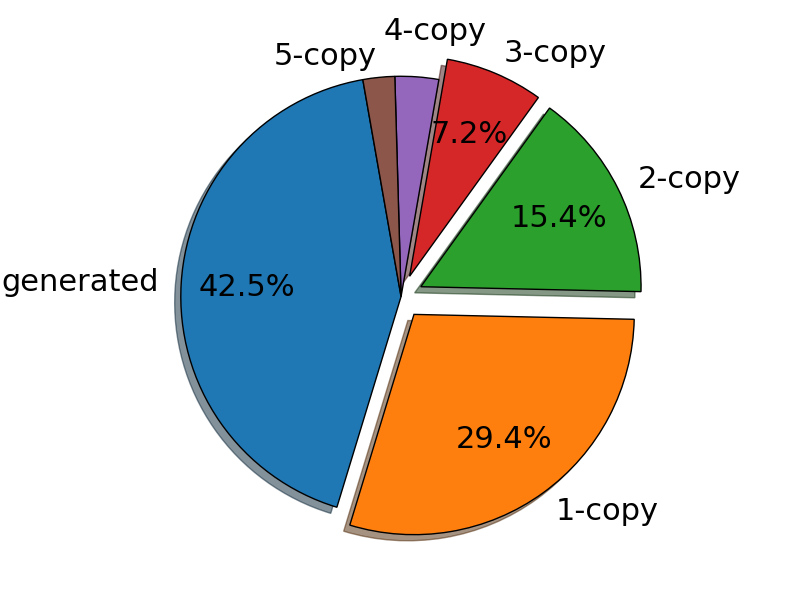}
	\caption{\label{fig:summDataStat} 
		Percentage of generated and copied words in sentence summarization training data.
	}
\end{figure}

\begin{figure}[htb]
	\centering
	\includegraphics[scale=0.15]{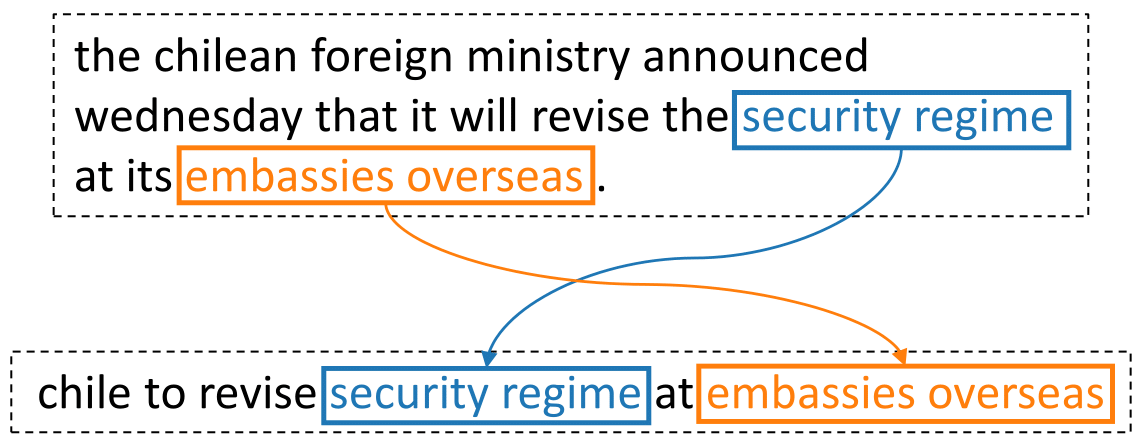}
	\caption{\label{fig:copyIntroExample} An example of sequential copying in abstractive sentence summarization task.}
\end{figure}

We further observe that the copied words usually form a continuous chunk of the output, exhibiting a ``sequential copying'' phenomenon. For example, in the Gigaword dataset of abstractive sentence summarization task, about 57.7\% words are copied from the input as indicated in Figure \ref{fig:summDataStat}.
Moreover, the copied words in multi-word span account for 28.1\%, which is also very common.
For example, in Figure \ref{fig:copyIntroExample}, there are two copied bi-grams in the output summary.
Similar phenomenon has also been observed in question generation task.

However, previous methods fall into one paradigm, which we call ``single word copy''.
At each decoding time step, the models still follow the ``word by word'' style to make separate decisions of whether to copy.
Therefore, this ``single word copy'' paradigm may introduce errors due to these separate decisions.
For example, the words in a phrase should be copied consecutively from the input sentence, but these separate decisions cannot guarantee to achieve this.
This may cause that some unrelated words appears unexpectedly in the middle of the phrase, or the phrase is not copied \red{completely} with some words missed.
Therefore, we argue that tasks such as abstractive sentence summarization and question generation can benefit from sequential copying considering the intrinsic nature of these tasks and datasets.

In this paper, we propose a novel copying framework, \textbf{Seq}uential \textbf{Copy}ing \textbf{Net}works (\ourModel{}), to extend the vanilla seq2seq framework.
\ourModel{} is intended to learn not only the ``single word copy'' behavior, but also the ``sequence copy'' operation as mentioned above.
We design a span extractor for the decoder so it can make ``sequence copy'' actions during decoding.
Specifically, \ourModel{} consists of three main components, an RNN based sentence encoder, an attention-equipped decoder, and the newly designed copying module.
We follow previous works to use the bidirectional RNN as the sentence encoder, and the decoder also employs an RNN with attention mechanism \cite{bahdanau2014neural}.
To achieve the sequential copying mechanism, the copying module is integrated with the decoder to make decisions during decoding.

The sequential copying module in \ourModel{} contains three main components, namely, the copy switch gate network, the pointer network and the copy state transducer.
The copy switch gate network is used to make decisions of whether to copy according to the current decoding states.
Its output is not a binary value, but a scalar \red{range} in [0, 1], which is the probability of choosing to copy.
The pointer network is then used to extract a span from the input sentence.
We maintains a copying state in the copying module so that the pointer network can make predictions based on it.
In detail, the pointer network predicts the start and end positions of the span.
The start position is predicted using the start copying state.
Then the copy state transducer will update the copying state so that the pointer network can predict the end position.
This transduction process is made by an RNN so that it can remember related information such as the start position, and guide the pointer to the corresponding end copying position.

We conduct experiments on abstractive sentence summarization and question generation tasks to verify the effectiveness of \ourModel{}.
On both tasks, \ourModel{} outperforms the baseline models and the case study show that it can copy meaningful spans.

\section{Sequential Copying Networks}

\begin{figure*}[htb]
	\centering
	\includegraphics[scale=0.25]{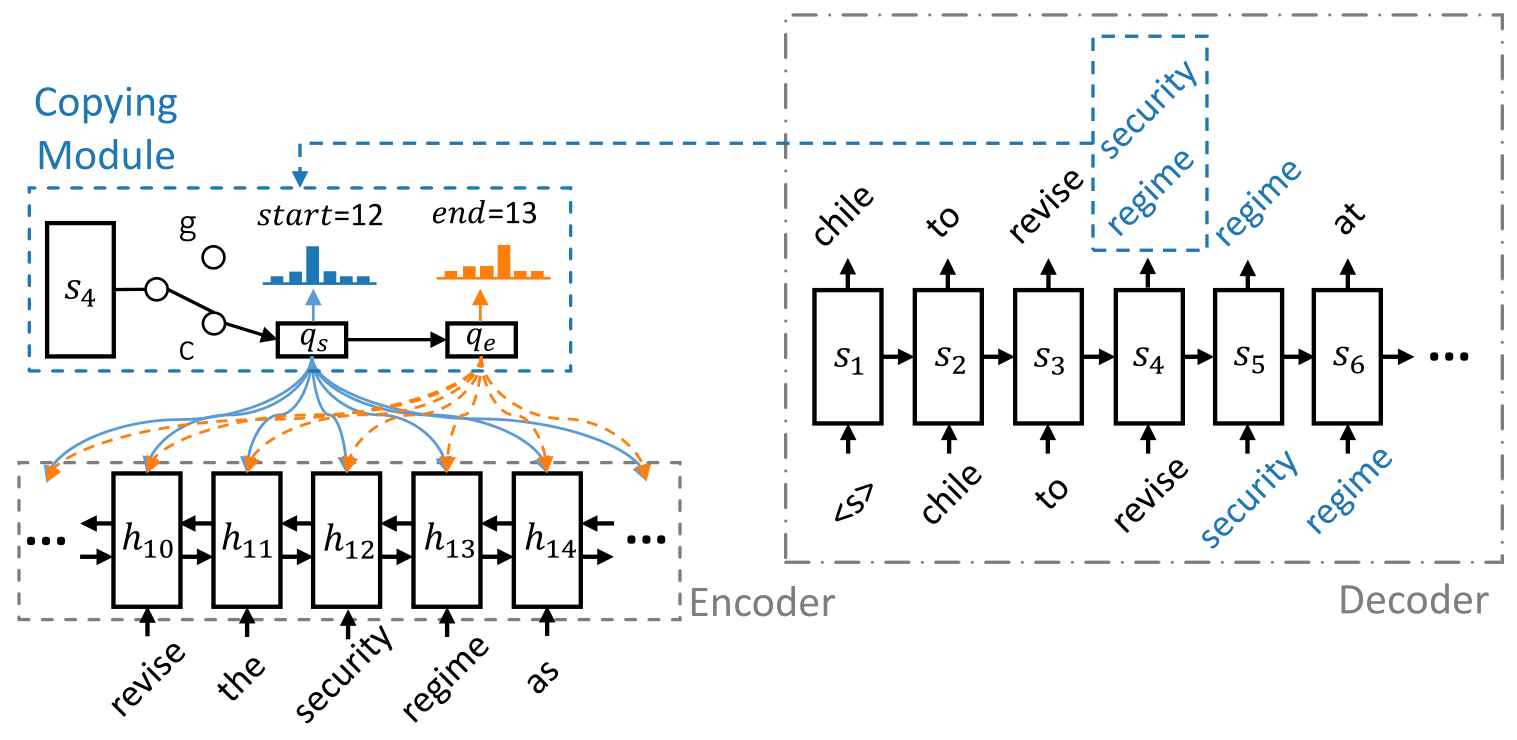}
	\caption{\label{fig:model} The overview diagram of \ourModel{}. For simplicity, we omit some units and connections. The copying process of the sequence ``security regime'' is magnified as indicated in the copying module part.}
\end{figure*}

As shown in Figure \ref{fig:model}, our \ourModel{} consists of three main components, namely the encoder, the copying module and the decoder.
Like in vanilla seq2seq frameworks, the encoder leverages two Gated Recurrent Unit (GRU) \cite{cho-EtAl:2014:EMNLP2014} to read the input words, and the decoder is modeled with GRU with attention mechanism.
The copying module consists of a copy switch gate network, a pointer network and a recurrent copy state transducer.
At each decoding time step, the copying module will make a decision of whether to copy or generate.
If it decides to copy, the pointer network and the copy state transducer will cooperate to copy a sub-span from the input sentence by predicting the start and end positions of it.
After the copying action, if the copied sequence contains more than one word, the decoder will apply ``Copy Run'' to update its states accordingly.

\subsection{Encoder}
The role of the sentence encoder is to read the input sentence and construct the basic sentence representation.
Here we employ a bidirectional GRU (BiGRU) as the recurrent unit, where GRU is defined as:
\begin{empheq}{align}
	z_i &= \sigma(\textbf{W}_z[x_i,h_{i-1}]) \\
	r_i &= \sigma(\textbf{W}_r[x_i,h_{i-1}]) \\
	\widetilde{h}_i &= \tanh(\textbf{W}_h[x_i,r_i \odot h_{i-1}]) \\
	h_i &= (1-z_i)\odot h_{i-1} + z_i \odot \widetilde{h}_i 
\end{empheq}
where $ \textbf{W}_z $, $ \textbf{W}_r $ and $ \textbf{W}_h $ are weight matrices.

The BiGRU consists of a forward GRU and a backward GRU.
The forward GRU reads the input sentence word embeddings from left to right and gets a sequence of hidden states, $ (\vec{h}_{1}, \vec{h}_{2}, \dots, \vec{h}_{n})  $.
The backward GRU reads the input sentence embeddings reversely, from right to left, and results in another sequence of hidden states, $ (\cev{h}_{1}, \cev{h}_{2}, \dots, \cev{h}_{n}) $:
\begin{empheq}{align}
	\vec{h}_{i} =\text{GRU}(x_{i}, \vec{h}_{i-1})\\
	\cev{h}_{i} = \text{GRU}(x_{i}, \cev{h}_{i+1})
\end{empheq}

The initial states of the BiGRU are set to zero vectors, i.e., $ \vec{h}_{1} = 0 $ and $ \cev{h}_{n} = 0 $.
After reading the sentence, the forward and backward hidden states are concatenated, i.e., $ h_{i} = [\vec{h}_{i} ; \cev{h}_{i}] $, to get the basic sentence representation.

\subsection{Sequential Copying Mechanism}

To model the sequential copying mechanism, \ourModel{} needs three key abilities:
a) at decoding time step $ t $, the model needs to decide whether to copy or not;
b) if the model decides to copy, it will need to select a sub-span from the input;
c) the decoder should switch between the generate mode and copy mode smoothly.
To enable our \ourModel{} of the first two functions, we design the copying module as a three-part component, i.e., the copy switch gate network, the pointer network and the copy state transducer.
The last ability is enabled by Copy Run method, which is described in the next section.
The copy switch gate network decides whether to copy during decoding.
If the model goes to generate mode, then it will generate next words as same as the vanilla attention-base seq2seq model.
If the model choose to copy, the pointer network will predict a sub-span.

At each time-step $ t $, the decoder GRU holds its previous hidden state $ s_{t-1} $, the previous output word $ y_{t-1} $ and the previous 
context vector $ c_{t-1} $. With these previous states, the decoder GRU updates its states as given by formula \ref{eq:decUpdate} .
To initialize the GRU hidden state, we use a linear layer with the last backward encoder hidden state $ \cev{h}_{1} $ as input:
\begin{empheq}{align}
\label{eq:decUpdate}s_{t} &= \text{GRU}(y_{t-1}, c_{t - 1}, s_{t-1})\\
s_{0} &= \tanh (\mathbf{W}_{d}\cev{h}_{1} + b)
\end{empheq}

With the new decoder hidden state $ s_{t} $, the context vector $ c_{t} $ for current time step $ t $ is computed through the concatenate attention mechanism \cite{luong-pham-manning:2015:EMNLP}, which matches the current decoder state $ s_{t} $ with each encoder hidden state $ h_{i} $ to get an importance score. The importance scores are then normalized to get the current context vector by weighted sum:
\begin{empheq}{align}
\label{eq:att10}e_{t,i} &= v_{a}^{\top}\tanh(\mathbf{W}_{a}s_{t} + \mathbf{U}_{a}h_{i})\\
\label{eq:att20}\alpha_{t,i} &= \frac{\exp (e_{t,i})}{\sum_{i=1}^{n}\exp (e_{t,i})}\\
\label{eq:att30}c_{t} &= \sum_{i = 1}^{n} \alpha_{t,i}h_{i}
\end{empheq}
where $ \mathbf{W}_{a} $ and $ \mathbf{U}_{a} $ are learnable parameters.

We then construct a new state vector and name it as decoder memory vector $ m_{t} $, which is the concatenation of the embedding of previous output word $ y_{t-1} $,  the decoder GRU hidden vector $ s_{t} $ and the current context vector $ c_{t} $:
\begin{equation}
m_{t} = \left[
\begin{matrix}
y_{t-1}\\
s_{t}\\
c_{t}
\end{matrix}
\right]
\end{equation}

In \ourModel{}, the decoder memory vector $ m_{t} $ plays an important role.
In the copying module, the copy switch gate network makes decisions based on $ m_{t} $.
Specifically, the copy switch gate network ($ \mathcal{G} $) is a Multilayer Perceptron (MLP) with two hidden layers:
\begin{equation}
\label{eq:gate}\mathcal{G}(x) = \sigma \left( \mathbf{W_{2}} (\tanh(\mathbf{W_{1}}x + b_{1})) + b_{2} \right)
\end{equation}
where $ \mathbf{W_{1}} $, $ \mathbf{W_{2}} $, $ b_{1} $ and $ b_{2} $ are learnable parameters.
The activation function of the first hidden layer is hyperbolic tangent (tanh).
To produce a probability of whether to copy, we use the sigmoid function ($ \sigma(\cdot) $ in Equation \ref{eq:gate}) as the activation function of the last hidden layer.
The copy probability $ p_{c} $ and generate probability $ p_{g} $ are defined as:
\begin{empheq}{align}
p_{c} &= \mathcal{G}(m_{t})\\
p_{g} & = 1 - p_{c}
\end{empheq}

\subsubsection{Generate Mode}
If the copy switch gate network decides to generate, \ourModel{} will generate the next word using the decoder memory vector $ m_{t} $.
The decoder first generates a readout state $ r_{t} $ and then pass it through a maxout hidden layer \cite{goodfellow2013maxout} to predict the next word with a softmax layer over the decoder vocabulary.
\begin{empheq}{align}
r_{t} &= \mathbf{W}_{r}w_{y-1} + \mathbf{U}_{r}c_{t} + \mathbf{V}_{r}s_{t}\\
\label{eq:maxout}r_{t}' &= [\max\{r_{t, 2j-1}, r_{t, 2j}\}]^{\top}_{j = 1,\dots, d}\\
p(y_{t} &\vert y_{<t}) = \text{softmax}(\mathbf{W}_{o}r_{t}')
\end{empheq}
where  $ \mathbf{W}_{r} $, $ \mathbf{U}_{r} $, $ \mathbf{V}_{r} $ and $ \mathbf{W}_{o} $ are weight matrices.
Readout state $ r_{t} $ is a $ 2d $-dimensional vector, and the maxout layer (Equation \ref{eq:maxout}) picks the max value for every two numbers in $ r_{t} $ and produces a d-dimensional maxout vector $ r_{t}' $.
We then apply a linear transformation on $ r_{t}' $ to get a target vocabulary size vector and predict the next word $ y_{t} $ with the  softmax operation.

\subsubsection{Copy Mode}
If the copy switch gate network decides to copy, then \ourModel{} uses its pointer network to predict a sub-span in the input sentence.
In detail, the pointer network makes two predictions, i.e., the start position prediction and the end position prediction.
The pointer network makes these predictions based on the decoder state.
Before deciding the start position, the copying module first generate a start query vector $ q_{s} $ using the decoder memory vector $ m_{t} $:
\begin{equation}
q_{s} = \tanh(\mathbf{W}_{s}m_{t} + b)
\end{equation}

Using the start query vector $ q_{s} $, the pointer network predicts the start position $ \text{copy}_{\text{s}} $ of the sub-span:
\begin{empheq}{align}
e_{s,i} &= v_{p}^{\top}\tanh(\mathbf{W}_{p}q_{s} + \mathbf{U}_{p}h_{i})\\
\alpha_{s,i} &= \frac{\exp (e_{s,i})}{\sum_{i=1}^{n}\exp (e_{s,i})}\\
\text{copy}_{\text{s}} &= \operatorname{arg\,max}_i \alpha_{s,i}\\
p_{\text{copy}\_{\text{s}}} & = \alpha_{s,\text{copy}_{\text{s}}}\\
c_{s} &= \sum_{i = 1}^{n} \alpha_{s,i}h_{i}
\end{empheq}
where  $ c_{s} $ is the copying context state vector, $ p_{\text{copy}\_{\text{s}}} $ is the probability of $ \text{copy}_{\text{s}} $ being the start copying position.

After predicting the start position, the copy state transducer will generate the end position query vector $ q_{e} $.
It is modeled with a single layer MLP and a GRU.
In detail, the MLP first produces an initial GRU hidden state $ cst $.
Then the GRU generates an end position query based on this hidden state $ cst $ and the copying context state $ c_{s} $:
\begin{empheq}{align}
cst &= \tanh(\mathbf{W}_{e}m_{t} + b)\\
q_{e} &= \text{GRU}(cst, c_{s})\\
e_{e,i} &= v_{p}^{\top}\tanh(\mathbf{W}_{p}q_{e} + \mathbf{U}_{p}h_{i})\\
\alpha_{e,i} &= \frac{\exp (e_{e,i})}{\sum_{i=1}^{n}\exp (e_{e,i})}\\
\text{copy}_{\text{e}} &= \operatorname{arg\,max}_i \alpha_{e,i}\\
p_{\text{copy}\_{\text{e}}} &= \alpha_{e,\text{copy}_{\text{e}}}
\end{empheq}
where $ \text{copy}_{\text{e}} $ is the end position and $ p_{\text{copy}\_{\text{e}}} $ is the probability of $ \text{copy}_{\text{e}} $ being the end copying position given the condition that $ \text{copy}_{\text{s}} $ is the start position.

After predicting the start and end positions, we can calculate the probability of copying this sub-span:
\begin{equation}
\begin{split}
p(\text{copy\_mode}, y_{t} = (x_{\text{copy}_{\text{s}}}, \dots, x_{\text{copy}_{\text{e}}} ) \vert y_{<t}) \\= p_{c} * p_{\text{copy}\_{\text{s}}} * p_{\text{copy}\_{\text{e}}}
\end{split}
\end{equation}

\subsection{Copy Run for Multi-word Span}
\label{sec:copyrun}
Since \ourModel{} may choose to copy a long sequence (copy mode) or generate a word from target vocabulary (generate mode), the decoder should adapt itself to smoothly switching between these two modes.
Suppose the decoder just copies a five-word span and the next word is generated, 
if we just skip these five copied words, the decoder RNN cannot remember the previous words when generating the next word.
This may lead to ill decoder RNN hidden states and poor performance of the sentence fluency and quality.
To solve this problem, we make the decoder GRU keep its state updated when it copies a long sequence.
Inspired by \citet{tang2016neural}, we propose a method, named Copy Run, to solve this problem in both training and testing phases.

During training, Copy Run solves this problem by keeping the decoder running over all output words.
By doing this, the decoder can learn from a complete sequence so the lack of decoder states update problem can be avoided.

During testing, Copy Run maintains the decoder GRU states by running over the copied words.
According to our model architecture, \ourModel{} decoder consumes the last generated word $ y_{t-1} $ and GRU state $ s_{t-1} $ at time step $ t $.
Therefore, the Copy Run only needs to be applied if the decoder copies a sequence whose length is larger than 1.
For a copied sequence with length $ l \geq 2 $, the Copy Run updates the decoder states for $ l - 1 $ times.

For example in Figure \ref{fig:model}, at time step $ 4 $, the decoder decides to copy a two-word ($ l = 2 $) sequence $ (x_{12}, x_{13}) $.
Using Copy Run, the decoder will feed the first $ l - 1 $ words into the GRU and generate a pseudo state $ s_{5} $.
Specifically, after the copying of $ (x_{12}, x_{13}) $, the copy run will execute the decoder as mentioned in Equation \ref{eq:decUpdate}, \ref{eq:att10}, \ref{eq:att20} and \ref{eq:att30} to update the GRU state $ s_{t} $ and the context vector $ c_{t} $, to prepare for the next time step.
For the rare words in copied sequence, the copy run will instead feed the embedding of \unk{} to update the decoder GRU states.

\subsection{Objective Function}
Given a training dataset with $ n $ input-output pairs $ \mathcal{D} = \lbrace(x^{(1)}, y^{(1)}), \dots, (x^{(n)}, y^{(n)})\rbrace $, where the $ k $-th pair is $ (x^{(k)}, y^{(k)}) = \left((x_{1}^{(k)}, \dots, x_{T_{x}}^{(k)} ),  (y_{1}^{(k)}, \dots, y_{T_{y}}^{(k)} )\right) $ and $ T_{x} $ and $ T_{y} $ are the lengths of $ x^{(k)} $ and $ y^{(k)} $ respectively.
For the  $ k $-th training instance $ (x^{(k)}, y^{(k)}) $, the set $ C_{k} $ contains the copied spans in $ y^{(k)} $.
Our training objective is to minimize the negative log likelihood loss $ \mathcal{L} $ with respect to the learnable model parameter $ \theta $:
\begin{equation}
\begin{split}
\mathcal{L}= -\frac{1}{n}\sum_{i=1}^{n}(\sum_{t=1}^{T_{y}}\log p_{g}p(y_{t}) + \sum_{span\in C_{k}} \log p_{c}p_{\text{start}}p_{\text{end}})
\end{split}
\end{equation}

\subsection{Beam Search}
Beam search is a common practical searching strategy to improve results for many tasks such as machine translation and dependency parsing.
We report both greedy search and beam search result in the experiments.
During beam search, we normalize the score of a beam path with its length.
For vanilla seq2seq models, this length equals to the number of generated words.
\red{For \ourModel{}, the length is defined as the summation of generated words and copied sequences.
	Taking the output in Figure \ref{fig:copyIntroExample} as an example, 
	the output sentence is `` chile to revise [security regime] at [embassies overseas]'', the length normalization term is 6, which means 4 generated words and 2 copied sequences.}
We empirically set beam size to 8 in our experiments.

\section{Experiments}
We conduct experiments on the abstractive sentence summarization and question generation tasks to demonstrate the effectiveness of \ourModel{}.

\subsection{Abstractive Sentence Summarization}
Sentence summarization aims to shorten a given sentence and produce a brief summary of it, and the models can be roughly categorized into extractive and abstractive methods.
Extractive methods select words from given inputs to form final summary sentence, while abstractive methods generate output summary after reading the input.
These two methods have their merits, the extractive methods use exactly the same words so the summary sentence is \red{accurate}, the abstractive methods can perform paraphrasing so the output can be more \red{diverse}.
After analyzing the dataset, we found that copying a sequence from input sentence happens in 57.5\% sentence-summary pairs (as shown in Figure \ref{fig:summDataStat}).
Therefore, we conduct experiments on this task.

\subsubsection{Dataset}

We conduct abstractive sentence summarization experiment on English Gigaword dataset, as mentioned in \citet{rush-chopra-weston:2015:EMNLP}.
The parallel corpus is produced by pairing the first sentence and the headline in the news article with some heuristic rules.
We modify the script released by \citet{rush-chopra-weston:2015:EMNLP} to pre-process and extract the training and development datasets.
We obtain the test set used by \citet{rush-chopra-weston:2015:EMNLP}.

However, like previous works mentioned \cite{chopra-auli-rush:2016:N16-1,zhou-EtAl:2017:Long}, there are many invalid lines in it and the scores reported on it cannot fully demonstrate the performance of the models.
So we acquired the test set sampled by \citet{zhou-EtAl:2017:Long}.
But we find that this test set is processed similar to \citet{rush-chopra-weston:2015:EMNLP}, that the rare words and numbers have already been replaced by \unk{} and \# symbol.
On test set like this, the copying methods can barely work since the unknown words and numbers in the references are already replaced, which is as shown in the experimental results.
Therefore, we further sample and release\footnote{We release the preprocessing script and this test set at \\ \url{http://res.qyzhou.me}  } a new test set, in which the sentence-summary pairs are remained untouched.
During training, we set the maximum copying length to 5.

\subsubsection{Baseline}
We compare \ourModel{} with the following baselines on abstractive sentence summarization task:
\begin{description}
	\item[ABS] \citet{rush-chopra-weston:2015:EMNLP} use an attentive CNN encoder and NNLM decoder for this task.
	\item[ABS+] Based on ABS model, \citet{rush-chopra-weston:2015:EMNLP} further tune ABS using DUC 2003 dataset.
	\item[RAS-Elman] As an extension of ABS, \citet{chopra-auli-rush:2016:N16-1} use a convolutional attention-based encoder and RNN decoder, which outperforms the ABS model.
	\item[Feats2s] \citet{nallapatiabstractive} use a full RNN sequence-to-sequence encoder-decoder model and add some features to enhance the encoder, such as POS tag, NER, and so on.
	\item[Luong-NMT] Neural machine translation model of \citet{luong-pham-manning:2015:EMNLP} with two-layer
	encoder-decoder	implemented in \citet{chopra-auli-rush:2016:N16-1}. 
	\item[s2s+att] We also implement a sequence-to-sequence model with attention as our baseline and denote it as ``s2s+att''.
	\item[NMT + UNK\_PS (single copy)]  We implement the UNK pointer softmax (PS) proposed by \citet{gulcehre-EtAl:2016:P16-1}.
	\item[SEASS] Selective encoding model for abstractive sentence summarization proposed by \citet{zhou-EtAl:2017:Long}.
\end{description}

\subsubsection{Evaluation Metric}
We employ \textsc{Rouge} \cite{lin2004rouge} as our evaluation metric.
\textsc{Rouge} measures the quality of summary by computing overlapping lexical units, such as unigram, bigram, trigram, and longest common subsequence (LCS).
It becomes the standard evaluation metric for DUC shared tasks and popular for summarization evaluation.
Following previous work, we use \textsc{Rouge}-1 (unigram), \textsc{Rouge}-2 (bigram) and \textsc{Rouge}-L (LCS) as the evaluation metrics in the reported experimental results.

\subsubsection{Model Parameters and Training}
The input and output vocabularies are collected from the training data with omitting the words appearing less than 20 times, which have 67,171 and 36,444 word types respectively.
We set the word embedding size to 300 and all GRU hidden state sizes to 512.
We use dropout \cite{srivastava2014dropout} with probability $ p = 0.4 $.
We initialize model parameters randomly using a Gaussian distribution with Xavier scheme \cite{glorot2010understanding}.
We use Adam \cite{kingma2014adam} as our optimizing algorithm.
For the hyperparameters of Adam optimizer, we set the learning rate $ \alpha = 0.001 $, two momentum parameters $ \beta_{1} = 0.9 $ and $ \beta_{2} = 0.999 $ respectively, and $ \epsilon=10^{-8} $.
During training, we test the model performance (\textsc{Rouge}-2 F1) on development set for every 2,000 batches.
We use the learning rate decay method, which is to halve the Adam learning rate $ \alpha $ if the \textsc{Rouge}-2 F1 score drops for six consecutive tests on development set.
We also apply gradient clipping \cite{pascanu2013difficulty} with range $ [-5, 5] $ during training.
To both speed up the training and converge quickly, we use mini-batch size 64 by grid search.
During test, we do post-processing by replacing the \unk{} with the token that has the highest attention score.

\begin{table}[htbp]
	\begin{center}
		\begin{tabular}{llll}
			\toprule
			\bf Models & \bf RG-1 & \bf RG-2 & \bf RG-L \\ 
			\midrule
			ABS (beam)\otherpaper{} & 29.55 & 11.32  & 26.42 \\
			ABS+ (beam)\otherpaper{} & 29.76 & 11.88 & 26.96 \\
			Feats2s (beam)\otherpaper{} & 32.67 & 15.59 & 30.64 \\
			RAS-Elman (greedy)\otherpaper{} & 33.10 & 14.45& 30.25 \\
			RAS-Elman (beam)\otherpaper{} & 33.78 & 15.97 & 31.15 \\
			Luong-NMT (beam)\otherpaper{} & 33.10 & 14.45 & 30.71 \\

			s2s+att (greedy)   & 34.95  & 16.51 &  32.54 \\
			s2s+att (beam)  & 35.77 & 17.34  & 33.24 \\

			NMT + UNK\_PS (greedy)  & 34.97 & 16.51 &  32.53 \\
			NMT + UNK\_PS (beam)  & 35.67 & 17.44  & 33.19 \\

			\midrule
			\ourModel{} (greedy) & 35.33 & 16.66 & 32.90 \\
			\ourModel{} (beam) & \textbf{35.93}  &  \textbf{17.51}   &  \textbf{33.35} \\
			\bottomrule
		\end{tabular}
	\end{center}
	\caption{\label{tb:gw_fb-table} Full length \textsc{Rouge} F1 evaluation results on the English Gigaword test set used by \citet{rush-chopra-weston:2015:EMNLP}. RG in the Table denotes \textsc{Rouge}. Results with \otherpaper{} mark are taken from the corresponding papers.
	}
\end{table}

\begin{table*}[htbp]
	\begin{center}
		\begin{tabular}{lcccccc}
			\toprule
			& \multicolumn{3}{c}{ Test set in \citet{zhou-EtAl:2017:Long}} & \multicolumn{3}{c}{Our internal test set} \\
			\cmidrule(lr){2-4}
			\cmidrule(lr){5-7}
			\bf Models & \bf RG-1 & \bf RG-2 & \bf RG-L & \bf RG-1 & \bf RG-2 & \bf RG-L \\ 
			\midrule
			ABS\otherpaper{} & 37.41\significant{} & 15.87\significant{} & 34.70\significant{} & - & - & - \\
			s2s+att (greedy)  & 46.21  & 24.02 & 43.30 & 45.46 & 22.83 & 42.66 \\
			s2s+att (beam)    & 47.08 & 25.11 & 43.81 & 46.54 & 24.18 & 43.55 \\
			NMT + UNK\_PS (greedy)  & 45.64 & 23.38 &  42.67 & 45.21 & 23.01 & 42.38 \\
			NMT + UNK\_PS (beam)   & 47.05 & 24.82 & 43.87 & 46.52 & 24.41 & 43.58 \\
			
			SEASS (greedy)\otherpaper{}  & 45.27  & 22.88 & 42.20 & - & - & - \\
			SEASS (beam)\otherpaper{}  & 46.86 & 24.58 & 43.53 & - & - & -  \\
			\midrule
			\ourModel{} (greedy)  & 46.51 & 24.14  & 43.20 & 46.08 & 23.99 & 43.26 \\
			\ourModel{} (beam)  & \textbf{47.27} & 25.07 & \textbf{44.00} & \textbf{47.13} & \textbf{24.93} & \textbf{44.06} \\
			\bottomrule
		\end{tabular}
	\end{center}
	\caption{\label{tb:internal-gw} Full length \textsc{Rouge} F1 evaluation on English Gigaword test set of \citet{zhou-EtAl:2017:Long} and our internal test set. Results with \otherpaper{} mark are taken from the corresponding papers. The ABS baseline fails to run with the latest stable \red{Torch} framework, so we omit its performance on our internal test set.
	}
\end{table*}

\subsubsection{Results}

We use the official ROUGE script (version 1.5.5) \footnote{\url{http://www.berouge.com/}} to evaluate the summarization quality in our experiments. For English Gigaword test sets  the outputs have different lengths so we evaluate the system with F1 metric \footnote{The ROUGE evaluation option is the same as \citet{rush-chopra-weston:2015:EMNLP}, -m -n 2 -w 1.2}.

Table \ref{tb:gw_fb-table} and \ref{tb:internal-gw} give the performance in terms of \textsc{Rouge}-F1 of the \ourModel{} and the baseline models on three test sets, i.e., \citet{rush-chopra-weston:2015:EMNLP} test set, \citet{zhou-EtAl:2017:Long} test set and our internal test set.
As indicated in these tables, \ourModel{} performs better than all the baseline models.
As previously mentioned, the performance of copying methods performs comparably to the s2s+att baseline on test sets of \citet{rush-chopra-weston:2015:EMNLP} and \citet{zhou-EtAl:2017:Long}, since the rare words and numbers have already been replaced in the references.

Therefore, on the untouched test set, which is more likely in the real application scenarios, the copying methods performs better than the vanilla seq2seq baseline.
\ourModel{} achieves 47.13 \textsc{Rouge}-1, 24.93 \textsc{Rouge}-2 and 44.06 \textsc{Rouge}-L F1 scores on this test set and performs better than the ``single word copy'' model.

\subsubsection{Case Study}

Table \ref{tbl:example_summ} gives three summarization examples generated by \ourModel{}.
These examples show that \ourModel{} can choose the correct span and generate meaningful summaries.
We also observe that the copying of named entities are very common as shown in the examples.
We can see that \ourModel{} can copy a sequence more completely while the ``single copy'' model occasionally misses some words such as the first and second examples.

\begin{table*}[htbp]
	\small
	\begin{center}
		\begin{tabular}{rp{0.78\textwidth}}
			\toprule
			\textbf{Input:} & david ortiz homered and scored three times , including the go-ahead run in the eighth inning , as the boston \hl{\textit{red sox}} beat the toronto \hl{\textit{blue jays 10-9}} in the american league on tuesday .\\
			\textbf{Reference:} & david ortiz helps red sox beat blue jays 10-9\\
			\textbf{SingleCopy:} & ortiz homers as red sox beat blue jays\\
			\textbf{\ourModel{}:} & [\hl{\textit{red sox}}] beat [\hl{\textit{blue jays 10-9}}]\\
			
			\midrule
			\textbf{Input:} & \hl{\textit{guyana 's president cheddi jagan}} , a long-time marxist turned free - marketeer , died here early thursday , an embassy spokeswoman said . he was 78 . \\
			\textbf{Reference:} & guyana 's president cheddi jagan marxist turned marketeer dies at 78\\
			\textbf{SingleCopy:} & guyana 's president jagan dies at 78\\
			\textbf{\ourModel{}:} & [\hl{\textit{guyana 's president cheddi jagan}}] dies at 78\\
			
			\midrule
			\textbf{Input:} & china topped myanmar 's marine \hl{\textit{product exporting countries annually}} in the past decade among over 40 's , the local voice weekly quoted the marine products producers and exporters association as reporting sunday .\\
			\textbf{Reference:} & china tops myanmars marine product exporting countries in past\\
			\textbf{SingleCopy:} & china tops myanmar 's marine product export\\
			\textbf{\ourModel{}:} & china tops myanmar 's marine [\hl{\textit{product exporting countries annually}}]\\
			
			\bottomrule
		\end{tabular}
	\end{center}
	\caption{\label{tbl:example_summ}Examples of generated summaries.
	The highlighted italic words in brackets are copied as a sequence by \ourModel{}.}
\end{table*}

\subsection{Question Generation}
Automatic question generation  from natural language text aims to generate questions taking text as input, which has the potential value of education purpose \cite{heilman2011automatic}.
As the reverse task of question answering, question generation also has the potential for providing a large scale corpus of question-answer pairs \cite{zhou2017neural}.
In this task, we also found that sequential copying is frequent so it may benefit from this.

\subsubsection{Model}
Following \citet{zhou2017neural}, we change the encoder to a feature-rich encoder and combine it with the \ourModel{}.
The feature-rich encoder reads the sentence words and handcrafted features to produce a sequence of word-and-feature vectors.
In detail, \citet{zhou2017neural} use four features, namely answer position, word case, POS tag and NER tag.
We follow this work and change the encoder in \ourModel{} accordingly.

\subsubsection{Dataset and Evaluation Metric}
We use the Stanford Question Answering Dataset (SQuAD) \cite{rajpurkar-EtAl:2016:EMNLP2016} as our training data.
SQuAD is composed of more than 100K questions posed by crowd workers on 536 Wikipedia articles.
Following \cite{zhou2017neural}, we acquired their training, development and test sets, which contain 86,635, 8,965 and 8,964 triples respectively.
We also use the same Stanford CoreNLP v3.7.0 \cite{manning-EtAl:2014:P14-5} to annotate POS and NER tags in sentences with its default configuration and pre-trained models.
We evaluate the model using BLEU-4 score \cite{papineni2002bleu}.
According to the results in \citet{zhou-EtAl:2017:Long} \red{there is correlation between BLEU score and human evaluation}, therefore we only report the model performance with BLEU metric.

\subsubsection{Baseline}
We compare \ourModel{} with the following baselines on question generation task:
\begin{description}
	\item[PCFG-Trans] The rule-based system modified on the code released by \citet{heilman2011automatic}.
	We modified the code so that it can generate question based on a given word span.
	\item[s2s+att] The seq2seq with attention mechanism without rich features as the baseline method.
	\item[NQG] The s2s+att baseline with the feature-rich encoder.
	\item[NQG+ (single copy)]  Based on NQG, pointing mechanism \cite{gulcehre-EtAl:2016:P16-1} is used to deal with rare words problem.
\end{description}

\begin{table}[htbp]
	\small
	\begin{center}
		\begin{tabular}{lcc}
			\toprule
			\textbf{Model} &  Dev set & Test set  \\
			\midrule
			PCFG-Trans\otherpaper{} & 9.28 & 9.31 \\
			s2s+att\otherpaper{} & 3.01 & 3.06  \\
			NQG\otherpaper{} & 10.06  & 10.13  \\
			NQG+\otherpaper{} (single copy) & 12.30  & 12.18 \\
			\midrule
			\ourModel{} & \textbf{13.13} & \textbf{13.02} \\
			\bottomrule
		\end{tabular}
	\end{center}
	\caption{\label{tbl:bleu_qg} BLEU-4 evaluation results. Models with \otherpaper{} mark are taken from the corresponding papers.}
\end{table}

\subsubsection{Results}
We report BLEU-4 scores on both development and test sets in Table \ref{tbl:bleu_qg}.
Note that the s2s+att baseline performs poorly since it does know the answer position information.
\ourModel{} outperforms the baseline models with 13.02 BLEU-4 score on test set and achieves the state-of-the-art result on this dataset.

\section{Related Work}

\subsection{Sequence-to-Sequence Learning}
\citet{sutskever2014sequence} propose sequence-to-sequence framework with Long Short-Term Memory (LSTM). It uses LSTM to encode the input words to a single hidden vector. Another LSTM is used to decode this sentence meaning vector to produce the target sentence.
\citet{bahdanau2014neural} extend it by introducing attention mechanism to align source and target words.
The encoder produces a list hidden vectors instead of single vector so the decoder can dynamically attended to different encoded vectors. This brings huge improvements in many seq2seq learning tasks, such as NMT \cite{bahdanau2014neural,luong-pham-manning:2015:EMNLP}, syntactic parsing \cite{vinyals2015grammar} and abstractive text summarization \cite{rush-chopra-weston:2015:EMNLP}.

\subsection{Pointing / Copying Mechanism}

\citet{gu-EtAl:2016:P16-1} and \citet{gulcehre-EtAl:2016:P16-1} propose similar copying / pointing mechanisms.
\citet{gu-EtAl:2016:P16-1} propose to score the source words and target vocabulary words together. This method builds an extended vocabulary which is the union of target vocabulary, source words and \{$ {unk} $\}. Then they apply a softmax over this extended vocabulary to decide the next output from this extended vocabulary.
\citet{gulcehre-EtAl:2016:P16-1} use a pointer softmax to handle the \unk{} or rare words problem in seq2seq learning.
During decoding, they use a copying gate to predict the copying probability and use the attention mechanism to choose the copied word.
They tried UNK pointer and entity pointer, and show that UNK pointer is better than entity pointer for abstractive sentence summarization task.

\citet{Pointer_Networks} propose Pointer Network to model the pointing behavior.
The pointer network is modeled with RNN and the pointing score is calculated with attention mechanism.
\citet{wang2016machine} propose match-LSTM and leverage pointer networks for Stanford Question Answering competition \cite{rajpurkar-EtAl:2016:EMNLP2016}.
The match-LSTM first matches the words in the query and passage, and then the pointer network predicts  the start and end positions of a answer span.

\subsection{Incorporating Phrase Information in NMT}
Phrase-based machine translation performs very well among the SMT systems \cite{Chiang:2005:HPM:1219840.1219873,koehn2009statistical}.
Recently, much work has been done to incorporate phrase information in NMT to further boost translation quality.
\citet{tang2016neural}, \citet{wang2017translating}, and \citet{dahlmann2017neural} propose similar approaches which use NMT decoder to select phrases generated by SMT system.

\section{Conclusion}
In this paper, we propose Sequential Copying Networks (\ourModel{}) to model the sequential copying phenomenon in sequence-to-sequence generation.
It leverages the pointer networks as the prediction component to dynamically extract spans from input sentence during the decoding process.
Experiments on abstractive sentence summarization and question generation tasks show that \ourModel{} has ability of copying a sub-span from input sentence.
In the future, we will apply \ourModel{} to other tasks such as multi-turn dialog response generation.

\bibliographystyle{aaai}
\bibliography{aaai}

\begin{thebibliography}{}

\bibitem[\protect\citeauthoryear{Bahdanau, Cho, and
  Bengio}{2015}]{bahdanau2014neural}
Bahdanau, D.; Cho, K.; and Bengio, Y.
\newblock 2015.
\newblock Neural machine translation by jointly learning to align and
  translate.
\newblock In {\em Proceedings of 3rd ICLR}.

\bibitem[\protect\citeauthoryear{Chiang}{2005}]{Chiang:2005:HPM:1219840.1219873}
Chiang, D.
\newblock 2005.
\newblock A hierarchical phrase-based model for statistical machine
  translation.
\newblock In {\em Proceedings of ACL}, ACL '05,  263--270.
\newblock Stroudsburg, PA, USA: Association for Computational Linguistics.

\bibitem[\protect\citeauthoryear{Cho \bgroup et al\mbox.\egroup
  }{2014}]{cho-EtAl:2014:EMNLP2014}
Cho, K.; van Merrienboer, B.; Gulcehre, C.; Bahdanau, D.; Bougares, F.;
  Schwenk, H.; and Bengio, Y.
\newblock 2014.
\newblock Learning phrase representations using rnn encoder--decoder for
  statistical machine translation.
\newblock In {\em Proceedings of EMNLP},  1724--1734.
\newblock Doha, Qatar: Association for Computational Linguistics.

\bibitem[\protect\citeauthoryear{Chopra, Auli, and
  Rush}{2016}]{chopra-auli-rush:2016:N16-1}
Chopra, S.; Auli, M.; and Rush, A.~M.
\newblock 2016.
\newblock Abstractive sentence summarization with attentive recurrent neural
  networks.
\newblock In {\em Proceedings of NAACL},  93--98.
\newblock San Diego, California: Association for Computational Linguistics.

\bibitem[\protect\citeauthoryear{Dahlmann \bgroup et al\mbox.\egroup
  }{2017}]{dahlmann2017neural}
Dahlmann, L.; Matusov, E.; Petrushkov, P.; and Khadivi, S.
\newblock 2017.
\newblock Neural machine translation leveraging phrase-based models in a hybrid
  search.
\newblock {\em arXiv preprint arXiv:1708.03271}.

\bibitem[\protect\citeauthoryear{Glorot and
  Bengio}{2010}]{glorot2010understanding}
Glorot, X., and Bengio, Y.
\newblock 2010.
\newblock Understanding the difficulty of training deep feedforward neural
  networks.
\newblock In {\em Aistats}, volume~9,  249--256.

\bibitem[\protect\citeauthoryear{Goodfellow \bgroup et al\mbox.\egroup
  }{2013}]{goodfellow2013maxout}
Goodfellow, I.~J.; Warde-Farley, D.; Mirza, M.; Courville, A.~C.; and Bengio,
  Y.
\newblock 2013.
\newblock Maxout networks.
\newblock {\em ICML (3)} 28:1319--1327.

\bibitem[\protect\citeauthoryear{Gu \bgroup et al\mbox.\egroup
  }{2016}]{gu-EtAl:2016:P16-1}
Gu, J.; Lu, Z.; Li, H.; and Li, V.~O.
\newblock 2016.
\newblock Incorporating copying mechanism in sequence-to-sequence learning.
\newblock In {\em Proceedings of the 54th Annual Meeting of the Association for
  Computational Linguistics (Volume 1: Long Papers)},  1631--1640.
\newblock Berlin, Germany: Association for Computational Linguistics.

\bibitem[\protect\citeauthoryear{Gulcehre \bgroup et al\mbox.\egroup
  }{2016}]{gulcehre-EtAl:2016:P16-1}
Gulcehre, C.; Ahn, S.; Nallapati, R.; Zhou, B.; and Bengio, Y.
\newblock 2016.
\newblock Pointing the unknown words.
\newblock In {\em Proceedings of the 54th Annual Meeting of the Association for
  Computational Linguistics (Volume 1: Long Papers)},  140--149.
\newblock Berlin, Germany: Association for Computational Linguistics.

\bibitem[\protect\citeauthoryear{Heilman}{2011}]{heilman2011automatic}
Heilman, M.
\newblock 2011.
\newblock {\em Automatic factual question generation from text}.
\newblock Ph.D. Dissertation, Carnegie Mellon University.

\bibitem[\protect\citeauthoryear{Kingma and Ba}{2015}]{kingma2014adam}
Kingma, D., and Ba, J.
\newblock 2015.
\newblock Adam: A method for stochastic optimization.
\newblock In {\em Proceedings of 3rd ICLR}.

\bibitem[\protect\citeauthoryear{Koehn}{2009}]{koehn2009statistical}
Koehn, P.
\newblock 2009.
\newblock {\em Statistical machine translation}.
\newblock Cambridge University Press.

\bibitem[\protect\citeauthoryear{Lin}{2004}]{lin2004rouge}
Lin, C.-Y.
\newblock 2004.
\newblock Rouge: A package for automatic evaluation of summaries.
\newblock In {\em Text summarization branches out: Proceedings of the ACL-04
  workshop}, volume~8.

\bibitem[\protect\citeauthoryear{Luong \bgroup et al\mbox.\egroup
  }{2015}]{luong-EtAl:2015:ACL-IJCNLP}
Luong, T.; Sutskever, I.; Le, Q.; Vinyals, O.; and Zaremba, W.
\newblock 2015.
\newblock Addressing the rare word problem in neural machine translation.
\newblock In {\em Proceedings of ACL},  11--19.

\bibitem[\protect\citeauthoryear{Luong, Pham, and
  Manning}{2015}]{luong-pham-manning:2015:EMNLP}
Luong, T.; Pham, H.; and Manning, C.~D.
\newblock 2015.
\newblock Effective approaches to attention-based neural machine translation.
\newblock In {\em Proceedings of EMNLP},  1412--1421.
\newblock Lisbon, Portugal: Association for Computational Linguistics.

\bibitem[\protect\citeauthoryear{Manning \bgroup et al\mbox.\egroup
  }{2014}]{manning-EtAl:2014:P14-5}
Manning, C.~D.; Surdeanu, M.; Bauer, J.; Finkel, J.; Bethard, S.~J.; and
  McClosky, D.
\newblock 2014.
\newblock The {Stanford} {CoreNLP} natural language processing toolkit.
\newblock In {\em Association for Computational Linguistics (ACL) System
  Demonstrations},  55--60.

\bibitem[\protect\citeauthoryear{Nallapati \bgroup et al\mbox.\egroup
  }{2016}]{nallapatiabstractive}
Nallapati, R.; Zhou, B.; glar Gul{\c{c}}ehre, {\c{C}}.; and Xiang, B.
\newblock 2016.
\newblock Abstractive text summarization using sequence-to-sequence rnns and
  beyond.
\newblock In {\em Proceedings of CoNLL}.

\bibitem[\protect\citeauthoryear{Papineni \bgroup et al\mbox.\egroup
  }{2002}]{papineni2002bleu}
Papineni, K.; Roukos, S.; Ward, T.; and Zhu, W.-J.
\newblock 2002.
\newblock Bleu: a method for automatic evaluation of machine translation.
\newblock In {\em Proceedings of ACL},  311--318.

\bibitem[\protect\citeauthoryear{Pascanu, Mikolov, and
  Bengio}{2013}]{pascanu2013difficulty}
Pascanu, R.; Mikolov, T.; and Bengio, Y.
\newblock 2013.
\newblock On the difficulty of training recurrent neural networks.
\newblock {\em ICML (3)} 28:1310--1318.

\bibitem[\protect\citeauthoryear{Rajpurkar \bgroup et al\mbox.\egroup
  }{2016}]{rajpurkar-EtAl:2016:EMNLP2016}
Rajpurkar, P.; Zhang, J.; Lopyrev, K.; and Liang, P.
\newblock 2016.
\newblock Squad: 100,000+ questions for machine comprehension of text.
\newblock In {\em Proceedings of EMNLP},  2383--2392.
\newblock Austin, Texas: Association for Computational Linguistics.

\bibitem[\protect\citeauthoryear{Rush, Chopra, and
  Weston}{2015}]{rush-chopra-weston:2015:EMNLP}
Rush, A.~M.; Chopra, S.; and Weston, J.
\newblock 2015.
\newblock A neural attention model for abstractive sentence summarization.
\newblock In {\em Proceedings of EMNLP},  379--389.
\newblock Lisbon, Portugal: Association for Computational Linguistics.

\bibitem[\protect\citeauthoryear{Srivastava \bgroup et al\mbox.\egroup
  }{2014}]{srivastava2014dropout}
Srivastava, N.; Hinton, G.~E.; Krizhevsky, A.; Sutskever, I.; and
  Salakhutdinov, R.
\newblock 2014.
\newblock Dropout: a simple way to prevent neural networks from overfitting.
\newblock {\em Journal of Machine Learning Research} 15(1):1929--1958.

\bibitem[\protect\citeauthoryear{Sutskever, Vinyals, and
  Le}{2014}]{sutskever2014sequence}
Sutskever, I.; Vinyals, O.; and Le, Q.~V.
\newblock 2014.
\newblock Sequence to sequence learning with neural networks.
\newblock In {\em Advances in neural information processing systems},
  3104--3112.

\bibitem[\protect\citeauthoryear{Tang \bgroup et al\mbox.\egroup
  }{2016}]{tang2016neural}
Tang, Y.; Meng, F.; Lu, Z.; Li, H.; and Yu, P.~L.
\newblock 2016.
\newblock Neural machine translation with external phrase memory.
\newblock {\em arXiv preprint arXiv:1606.01792}.

\bibitem[\protect\citeauthoryear{Vinyals and Le}{2015}]{vinyals2015neural}
Vinyals, O., and Le, Q.
\newblock 2015.
\newblock A neural conversational model.
\newblock {\em arXiv preprint arXiv:1506.05869}.

\bibitem[\protect\citeauthoryear{Vinyals \bgroup et al\mbox.\egroup
  }{2015}]{vinyals2015grammar}
Vinyals, O.; Kaiser, {\L}.; Koo, T.; Petrov, S.; Sutskever, I.; and Hinton, G.
\newblock 2015.
\newblock Grammar as a foreign language.
\newblock In {\em NIPS},  2773--2781.

\bibitem[\protect\citeauthoryear{Vinyals, Fortunato, and
  Jaitly}{2015}]{Pointer_Networks}
Vinyals, O.; Fortunato, M.; and Jaitly, N.
\newblock 2015.
\newblock Pointer networks.
\newblock In Cortes, C.; Lawrence, N.~D.; Lee, D.~D.; Sugiyama, M.; and
  Garnett, R., eds., {\em Advances in Neural Information Processing Systems
  28}. Curran Associates, Inc.
\newblock  2692--2700.

\bibitem[\protect\citeauthoryear{Wang and Jiang}{2016}]{wang2016machine}
Wang, S., and Jiang, J.
\newblock 2016.
\newblock Machine comprehension using match-lstm and answer pointer.
\newblock In {\em Proceedings of ICLR}.

\bibitem[\protect\citeauthoryear{Wang \bgroup et al\mbox.\egroup
  }{2017}]{wang2017translating}
Wang, X.; Tu, Z.; Xiong, D.; and Zhang, M.
\newblock 2017.
\newblock Translating phrases in neural machine translation.
\newblock {\em arXiv preprint arXiv:1708.01980}.

\bibitem[\protect\citeauthoryear{Zhou \bgroup et al\mbox.\egroup
  }{2017a}]{zhou2017neural}
Zhou, Q.; Yang, N.; Wei, F.; Tan, C.; Bao, H.; and Zhou, M.
\newblock 2017a.
\newblock Neural question generation from text: A preliminary study.
\newblock {\em arXiv preprint arXiv:1704.01792}.

\bibitem[\protect\citeauthoryear{Zhou \bgroup et al\mbox.\egroup
  }{2017b}]{zhou-EtAl:2017:Long}
Zhou, Q.; Yang, N.; Wei, F.; and Zhou, M.
\newblock 2017b.
\newblock Selective encoding for abstractive sentence summarization.
\newblock In {\em Proceedings of the 55th Annual Meeting of the Association for
  Computational Linguistics (Volume 1: Long Papers)},  1095--1104.
\newblock Vancouver, Canada: Association for Computational Linguistics.

\end{thebibliography}

\end{document}